\documentclass[pmlr]{jmlr}

\usepackage{longtable}
\usepackage{microtype}

 \usepackage{booktabs}
 
\usepackage[load-configurations=version-1]{siunitx} 

\makeatletter
\def\set@curr@file#1{\def\@curr@file{#1}} 
\makeatother

\definecolor{darkred}{RGB}{200, 50, 50}

\theorembodyfont{\upshape}
\theoremheaderfont{\scshape}
\theorempostheader{:}
\theoremsep{\newline}

\title [HD computing for feature selection]{Hyperdimensional computing encoding for feature selection
on the use case of epileptic seizure detection}

\author{\Name{Una Pale}
      \Email{una.pale@epfl.ch}\\ 
      \addr Embedded Systems Laboratory (ESL)\\
      Swiss Federal Institute of Technology Lausanne (EPFL)\\
      Lausanne, Switzerland 
      \AND
      \Name{Tomas Teijeiro}
      \Email{tomas.teijeiro@epfl.ch}\\ 
      \addr Embedded Systems Laboratory (ESL)\\
      Swiss Federal Institute of Technology Lausanne (EPFL)\\
      Lausanne, Switzerland 
      \AND
      \Name{David Atienza}
      \Email{david.atienza@epfl.ch}\\ 
      \addr Embedded Systems Laboratory (ESL)\\
      Swiss Federal Institute of Technology Lausanne (EPFL)\\
      Lausanne, Switzerland } 

\begin{document}

\maketitle

\begin{abstract}
The healthcare landscape is moving from the reactive interventions focused on symptoms treatment to a more proactive prevention, from one-size-fits-all to personalized medicine, and from centralized to distributed paradigms. Wearable IoT devices and novel algorithms for continuous monitoring are essential components of this transition. Hyperdimensional (HD) computing is an emerging ML paradigm inspired by neuroscience research with various aspects interesting for IoT devices and biomedical applications. Here we explore the not yet addressed topic of optimal encoding of spatio-temporal data, such as electroencephalogram (EEG) signals, and all information it entails to the HD vectors. Further, we demonstrate how the HD computing framework can be used to perform feature selection by choosing an adequate encoding. To the best of our knowledge, this is the first approach to perform feature selection using HD computing in the literature. As a result, we believe it can support the ML community to further foster the research in multiple directions related to feature and channels selection, as well as model interpretability. 

\end{abstract}

\section{Introduction}
Hyperdimensional (HD) computing is an emerging ML paradigm inspired by neuroscience research, pointing towards the power of computation with massively distributed and redundant systems. Instead of representing data as conventional numbers, representing/projecting it to thousands (or millions) of neurons/numbers brings new perspectives. More specifically, HD computing or vector-symbolic architectures represent data in the shape of vectors with high dimensions, \textit{hypervectors}, with usually more than 10000 values. This change in the data representation paradigm brings various advantages from a learning and hardware implementation perspective. From a learning perspective, it opens new paths for semi-supervised~\cite{imani_semihd_2019}, distributed~\cite{imani_framework_2019}, continuous online learning~\cite{moin_wearable_2021,benatti_online_2019}, or multi-centroid learning~\cite{pale_multi-centroid_2022}. From the hardware implementation perspective, possibilities for parallelization, open paths for designing efficient accelerators~\cite{imani_revisiting_2021} or in-memory computations~\cite{karunaratne_-memory_2020, karunaratne_energy_2021}. 
Its lower energy and memory requirements ~\cite{burrello_ensemble_2021,asgarinejad_detection_2020, imani_sparsehd_2019} makes it an attractive solution for learning on less powerful devices, such as wearable and IoT systems. 

Due to the advantages mentioned above, HD computing is drawing a lot of attention for biomedical applications, and has indeed been tested for electromyogram (EMG), gesture recognition~\cite{rahimi_hyperdimensional_2016}, EEG error-related potentials detection~\cite{rahimi_hyperdimensional_2020}, emotion recognition from GSR (galvanic-skin response) and electrocardiogram (ECG) and electroencephalogram (EEG)~\cite{chang_hyperdimensional_2019}, and epileptic seizure detection from EEG~\cite{burrello_laelaps_2019, asgarinejad_detection_2020} among others. 

EEG (iEEG) and EMG are spatio-temporal, noisy and non-stationary data, whose efficient encoding to HD vectors poses an essential part of the quality of HD learning. However, how to encode channel and feature information has not yet been systematically explored in the literature. In most of the existing literature utilizing EMG or EEG data, only raw data or \textit{local-binary-patterns} (LBPs, ~\cite{kaya_1d-local_2014}, have been used as features encoded to vectors. Yet, similarly to standard ML approaches, the possibility to add more features can significantly improve the power of the models. In~\cite{pale_systematic_2021}, the authors showed that utilizing a wider variety of statistical, time, or frequency features with HD computing for epileptic seizure, the final detection score outperforms simple raw signal encoding. Thus, in this paper, we discuss possibilities on how to encode information about channels and various possible features one might need to extract from data, to capture all the relevant aspects.

As we shown later in this work, encoding can also enable analyzing performance, correlations, and learning capabilities of individual features. In~\cite{burrello_ensemble_2021} authors explored a similar idea using several individual HD classifiers for three different features. Analyzing various aspects of individual features is a crucial step for a) feature selection, as demonstrated in this paper, and b) for further interpretability of model decisions (which will not be discussed in the scope of this paper). Indeed, feature selection is crucial in designing wearable applications as it helps remove noisy and non-informative features while also directly leading to more lightweight models. To the best of our knowledge, a clear, straightforward methodology for feature selection using HD computing has not been so far presented in the literature. 

In this paper, we assess our approaches in the context of epileptic seizure detection. Epileptic seizures are a chronic neurological disorder characterized by the unpredictable occurrence of seizures that affects a significant portion of the world population (0.6 to 0.8\%)~\cite{mormann_seizure_2007}. Due to its high inter-patient variability, unpredictable nature and not yet completely understood origins, it still poses an open research question. Despite pharmacological treatments, one-third of patients still suffer from seizures~\cite{schmidt_evidence-based_2012} and are being subject to serious health risks and many restrictions in daily life. There is no yet available wearable device for epilepsy prediction, detection, or continuous monitoring for outpatient environments. HD computing has been utilized for epilepsy detection due to its attractive properties, and only with various improvements to standard HD learning, it can reach the performance of state-of-the-art algorithms~\cite{pale_exploration_2022}. Thus, further exploration on how to optimize HD computing and encoding for better performance is of high interest for epilepsy detection. 

Speaking more broadly, the healthcare landscape is moving from reactive disease intervention to proactive prevention~\cite{waldman_healthcare_2019}, as it is not only more cost-effective, but it usually leads to better quality of life~\cite{goetzel_prevention_2009}. At the same time it is shifting from one-size-fits-all to personalized medicine~\cite{chan_personalized_2011} and institution-centered to decentralized~\cite{puri_artificial_nodate}. For all these aspects, novel algorithms design and optimization for lightweight and wearable IoT devices are essential~\cite{tricoli_wearable_2017}. Thus, even though we focus our work on epilepsy detection, proposed approaches and results are applicable for more general continuous monitoring, early detection of diseases, and preventive healthcare. 






\subsection*{Generalizable Insights about ML in the Context of Healthcare}
 
This paper demonstrates how a novel HD computing approach can be used as an alternative to standard state-of-the-art machine learning (ML) approaches in the use case of epileptic seizure detection. We explore not yet addressed topic of optimal encoding of spatio-temporal data, such as EEG, and all information it entails to the HD vectors. Computational efficiency is critical for developing wearable devices for continuous monitoring of diseases such as epilepsy, making long battery lifetimes feasible and moving towards preventive healthcare. 
Further, we demonstrate an example of how we can utilize the HD computing framework to perform feature selection by choosing an adequate encoding. To the best of our knowledge, this is the first proposal of the feature selection using HD computing in the literature, and we trust it can foster research in the ML community for further development in this direction. 
From the clinical perspective, tools and models that can improve the interpretability of the models can lead not only to better service to the patients, but also to help as decision support for the doctors and caregivers.


\begin{figure}[t]
  \centering 
  \includegraphics[width=4in]{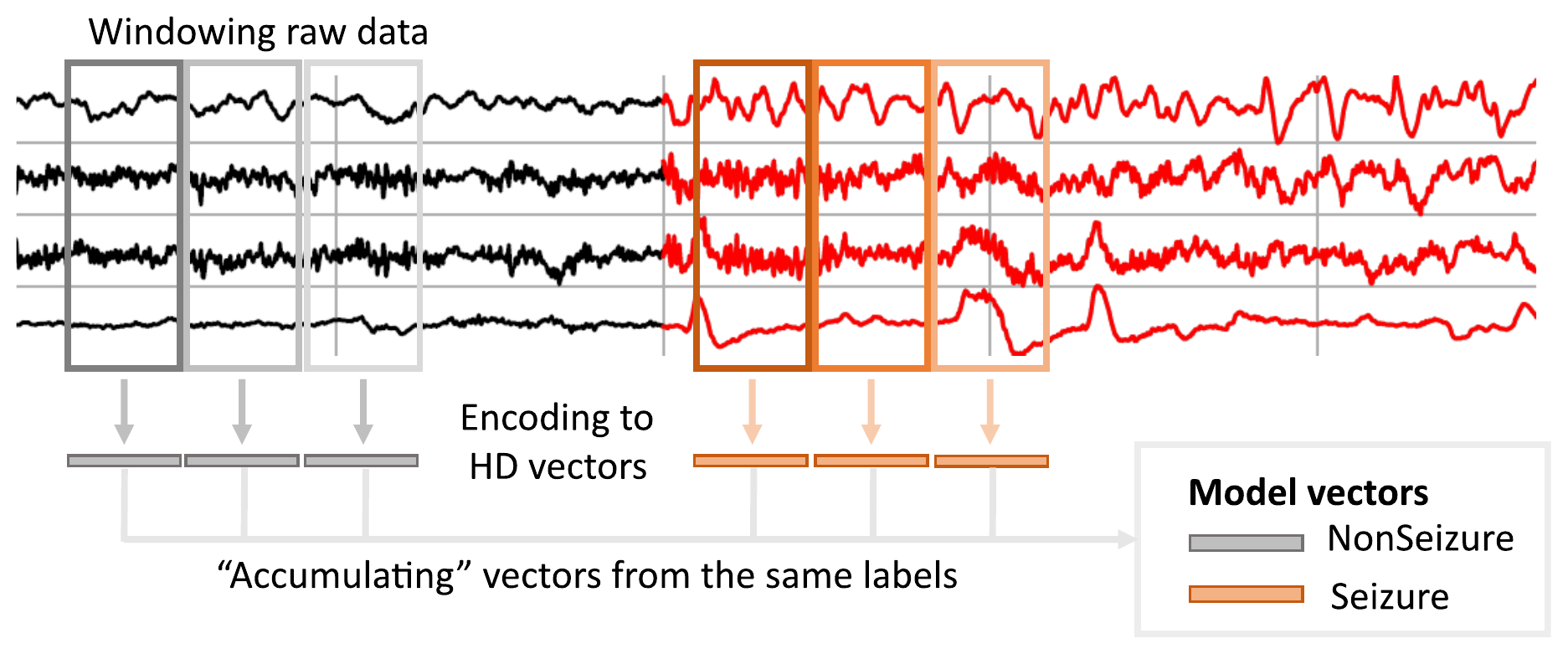} 
  \caption{Simple illustration showing the HD computing workflow and learning process from windows of seizure and non-seizure data.}
  \label{fig:HDworkflow} 
  \vspace{-8mm}
\end{figure}

\section{Related Work}

HD computing has been applied so far in several applications where spatio-temporal data such as EEG, iEEG~\cite{asgarinejad_detection_2020, burrello_ensemble_2021} or EMG~\cite{rahimi_hyperdimensional_2016, benatti_online_2019,moin_adaptive_2019, rahimi_hyperdimensional_2020} was utilized. The majority of works used raw signal values or short local-binary-patterns (LBPs) to describe the changes of the signal in time and map it directly to HD vectors~\cite{rahimi_hyperdimensional_2016, benatti_online_2019,moin_adaptive_2019, burrello_laelaps_2019, asgarinejad_detection_2020, rahimi_hyperdimensional_2020}. This enables combining channel and value (raw signal or LBP value) together and then further combining all samples belonging to the same class, similarly as illustrated on Fig.~\ref{fig:HDworkflow}. Some of them have also encoded time information between neighboring samples by utilizing vector permutation. Furthermore, in~\cite{karunaratne_energy_2021} authors propose energy-efficient in-memory encoding for this way of encoding spatio-temporal signals. This is an interesting approach in its simplicity as it use directly raw signal values. However, it is limited to only mapping one type of information to HD vectors (i.e., raw signal amplitude or/and signal change trends).

In~\cite{pale_systematic_2021} authors test, on a use-case of epilepsy seizure detection, how different feature types regularly used in non-HD ML classifiers for epilepsy detection perform in the HD computing framework. They show that, indeed, utilizing more different statistical, time, or frequency features outperforms simple raw signal encoding. 
Further, in recent~\cite{burrello_ensemble_2021}, authors extend their previous work~\cite{burrello_laelaps_2019} by adding mean amplitude and line length features to LPB values. They resolved the problem of encoding more features to HD vectors by having three independent classifiers, each with its own model vectors. Predictions are merged using a simple linear layer that gives a final prediction based on distances from class models of each feature. This is a promising approach that enables the performance comparison for the different features. 

\begin{figure}[t]
  \centering 
  \includegraphics[width=6in]{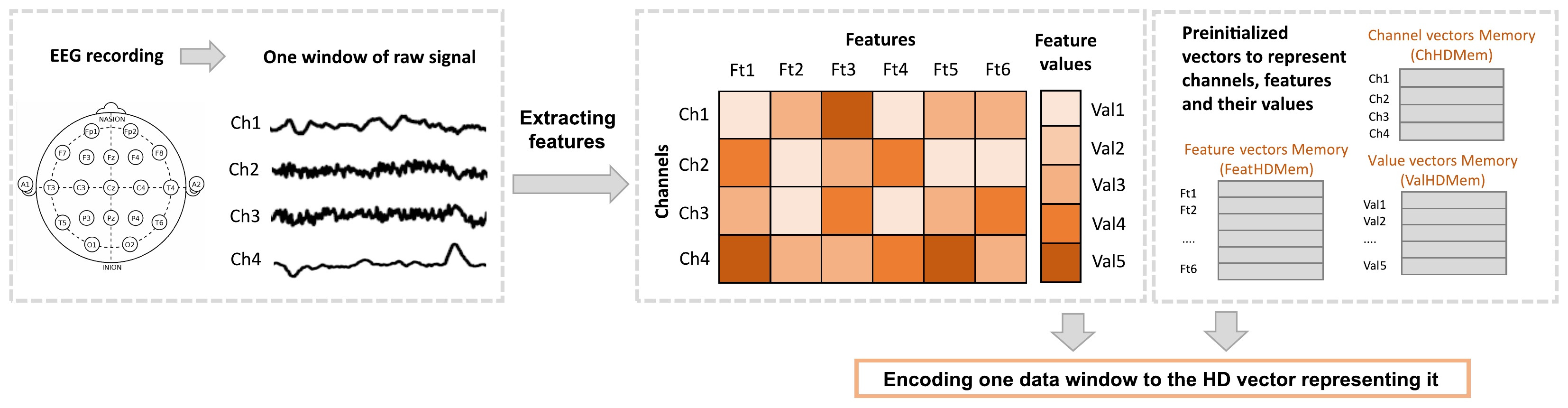} 
  \caption{Illustration of encoding one data window to HD vector representing it.}
  \label{fig:EncodingIllustration} 
  \vspace{-6mm}
\end{figure} 

In~\cite{ge_seizure_2021}, authors explored epileptic seizure detection using power-spectral features from iEEG data, and encountered similar problems as we point out here. They explored three different encoding approaches: 1) concatenating feature HD vectors to generate long HD vectors, 2) using multiple classifiers (one for each feature) and then integrating their predictions by majority voting, and 3) training one classifier using all features. In the last approach, vectors representing features and their values are combined in the first place, and then united with channel information. These approaches are highly interesting due to the broad spectrum of feature characteristics that are integrated. However, unfortunately, in~\cite{ge_seizure_2021} they were not systematically compared from performance, memory or complexity aspects. 




\section{Methods}
\subsection{Encoding spatio-temporal data for HD computing}

Typical spatio-temporal data, such as EMG or EEG, is 2D data. It consists of several channels positioned in different physical locations (which can have specific relations between them) and recorded during time frames of various lengths. After preprocessing the signals, features are typically extracted in time, using time windows shifted by some time step. As this is repeated for each of the channels, this leads to 3D data containing information on the feature, channel, and time, as illustrated in Fig.~\ref{fig:EncodingIllustration}. Within the HD computing workflow, this means that we need to define initial vectors that represent each of these entities: HD vectors representing each feature ($Feat_{ID}$ as in Fig.~\ref{fig:EncodingIllustration}), and vectors representing each possible value of features ($FeatCh_{VAL}$), and channels ($Ch_{ID}$). As feature values of various features might not have the same range and are not usually integer values, they need to be normalized and discretized to a specific number of bins. 

\begin{figure}[t]
  \centering 
  \includegraphics[width=6in]{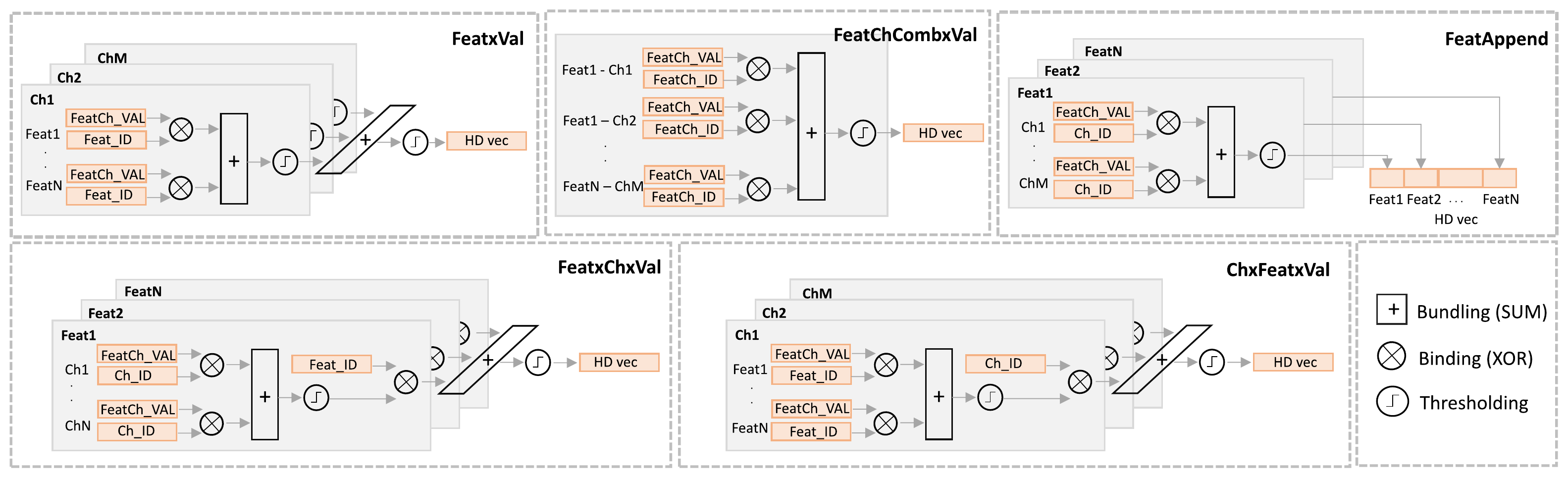} 
  \caption{Schematic of different possibilities to encode information about channels, features and their values to HD vectors. Five different encoding approaches tested in the paper are illustrated.}
  \label{fig:EncodingApproaches} 
  \vspace{-8mm}
\end{figure}

The question that we address in this work is how to encode all this information into HD vectors. Moreover, we assess the different possible methods from several perspectives, which include classification performance, memory and computational complexity. The last two metrics are very relevant in the design of the next generation of smart wearable systems. Therefore, in Fig.~\ref{fig:EncodingApproaches} we illustrate different possibilities considered in this paper to encode the time window of data to an HD vector. The time window contains information on all features and their values calculated from each of the channels. 

Although HD vectors can be binary (containing only 0 or 1), bipolar (containing -1 and 1), ternary (containing -1, 0 or 1), integer or even floating point, the most memory friendly and commonly used are binary HD vectors. Three basic arithmetic operations that are performed on the vectors are: 1)~bundling or bitwise summation, 2)~binding or bitwise XOR, and 3) thresholding to binarize vectors after summation. Bundling operation leads to a vector that is with high probability very similar to summed vectors, while on the other hand, binding leads to the vector that is orthogonal to the bound vectors. Thresholding is necessary to keep the final vector again in binary (bipolar or tertiary form) after summation. 

A typical approach in the literature dealing with simpler, non-spatial data is to bind feature vectors with feature value vectors and then bundle (and threshold) them. Translating this approach to EEG data leads to two options: 1)~$Feat\times Val$ and 2)~$ChFeatComb\times Val$. The difference between them is, as illustrated in Fig.~\ref{fig:EncodingApproaches} and formulated by (\ref{eq_FeatxVal}  and \ref{eq_ChFetaCombxVal}), that $Feat\times Val$ does not include information about channels but simply bundles (and normalizes) features ($Feat_{ID}$) and feature values from each channel ($FeatCh_{VAL}$). 
On the other side,  $ChFeatComb\times Val$ approach treats each feature and channel combination as an individual feature and gives it an independent HD vector ($FeatCh_{ID}$). This approach distinguishes between channels as formulated in (\ref{eq_ChFetaCombxVal}), but can lead to significantly large memory maps to store all initial HD vectors. This could happen in case data contains many channels (such as iEEG data) or when many features are extracted from each channel. 

As this can be problematic from a memory perspective viewpoint in the context of wearable medical devices, we propose other more EEG-inspired approaches: 3)~$Feat\times Ch\times Val$ and 4)~$Ch\times Feat\times Val$. Both approaches first initialize vectors for each channel ($Ch_{ID}$), each feature ($Feat_{ID}$), and possible feature values ($FeatCh_{VAL}$). The difference is, in the order of bundling information. For $Feat \times Ch \times Val$, as formulated by (\ref{eq_FeatxChxVal}), in the first step, channels ($Ch_{ID}$) and feature values on that channels ($FeatCh_{VAL}$) are bundled. Then, in the second step, the bundling with feature ID's ($Feat_{ID}$) follows. In $Ch \times Feat \times Val$ approach order of bundling is opposite, as formulated in (\ref{eq_ChxFeatxVal}).


\begin{equation}\label{eq_FeatxVal}
Feat \times Val = \Big \lfloor \sum_{i=1}^{numCh*numFeat} {Feat_{IDi} \oplus FeatCh_{VALi}}  \Big \rfloor
\end{equation}

\begin{equation}\label{eq_ChFetaCombxVal}
ChFeatComb \times Val =\Big \lfloor \sum_{i=1}^{numCh*numFeat} { FeatCh_{IDi} \oplus FeatCh_{VALi} }  \Big \rfloor
\end{equation}

\begin{equation}\label{eq_FeatxChxVal}
Feat \times Ch \times Val =\bigg \lfloor \sum_{i=1}^{numFeat} {Feat_{IDi} \oplus  \Big \lfloor \sum_{j=1}^{numCh} {Ch_{IDj} \oplus FeatCh_{VALij}} \Big \rfloor }  \bigg \rfloor
\end{equation}

\begin{equation}\label{eq_ChxFeatxVal}
Ch \times Feat \times Val =\bigg \lfloor \sum_{i=1}^{numCh} {Ch_{IDi} \oplus  \Big \lfloor \sum_{j=1}^{numFeat} {Feat_{IDj} \oplus FeatCh_{VALij}}  \Big \rfloor }  \bigg \rfloor
\end{equation}

The last approach, called $FeatAppend$, is designed with the goal to make it easier to interpret encoded vectors and determine which part of them comes from different features. In this approach, as illustrated in Fig.~\ref{fig:EncodingApproaches}, channels and feature values are bound, bundled, and thresholded to get a vector representing the encoded sub-vector for each feature. Instead of binding it with other feature sub-vectors as in $Feat \times Ch \times Val$ these vectors are appended one after another, as formulated in (\ref{eq_FeatAppend}). In order to get the same final vector dimension, initialized sub-vectors have, in this case, smaller dimensions.  
This encoding organization enables to analyze vectors for every single feature. For example, in this paper, we analyze what the separability of classes for each of the features is, as well as the predictions of individual features, and how confident they are with respect to other features. These measures are defined in next section and allow us to perform additional features selection steps. 

\begin{equation}\label{eq_FeatAppend}
FeatAppend = \Big \lfloor \sum_{j=1}^{numCh} {Ch_{IDj} \oplus Feat_{1}Ch_{VALj}}  \Big \rfloor . . .   \Big \lfloor \sum_{j=1}^{numCh} {Ch_{IDj} \oplus  Feat_{numFeat}Ch_{VALj}}  \Big \rfloor 
\vspace{-.5cm}
\end{equation}

\subsection{Feature selection } \label{sec:featureSelection} 

In the $FeatAppend$ approach, it is known which part of the final encoded vector comes from each feature, which enables analysis per individual feature. In this paper, we define and measure several metrics for each feature: 
\begin{itemize}
    \item \textit{Prediction}: Using only $d=D/numFeat$ bits corresponding to the feature of interest, we can determine the prediction for each sample using only that feature. A decision is made in the same way as when the whole vector is used; the label of the most similar class vector is given. As we use binary HD vectors, hamming distance is used as our similarity metric measure. 
    

    \item \textit{Feature certainty}: For each time moment, we can also quantify the certainty of each feature's label based on distances from both class vectors. The certainty is calculated as the difference between distances from two classes, divided by the average absolute distance for all features in the same time moment, as in formula (\ref{eq_probab}).
    
\begin{equation}\label{eq_probab}
C_f = \frac{|{dist}_{Sf} - {dist}_{NSf}|}{  \frac{1}{numFeat} \sum_{i=1}^{numFeat} {(|{dist}_{Si} - {dist}_{NSi}|)}} 
\end{equation}
    
    \item \textit{Correlation}: Based on the predictions of each feature in time, it is possible to measure the correlation between  predictions of different features and take it into account later for feature selection. 
    
    \item \textit{Class separability}: For binary HD vectors, separability ($S_{f}$) is measured as the relative hamming distance between class vectors ($HD_{S}, HD_{Ns}$) when using only the bits of the corresponding feature. This measurement is obtained using equation (\ref{eq_separability}). Based on the separability measure it is also possible to estimate the usefulness of each feature.
\begin{equation}\label{eq_separability}
S_f = hamming\Bigg( HD_{S}\bigg[\frac{f*D}{numFeat}:\frac{(f+1)*D}{numFeat} \bigg], HD_{NS}\bigg[\frac{f*D}{numFeat}:\frac{(f+1)*D}{numFeat} \bigg] \Bigg)
\end{equation}
    
\end{itemize}

\begin{figure}[t]
  \centering 
  \includegraphics[width=6in]{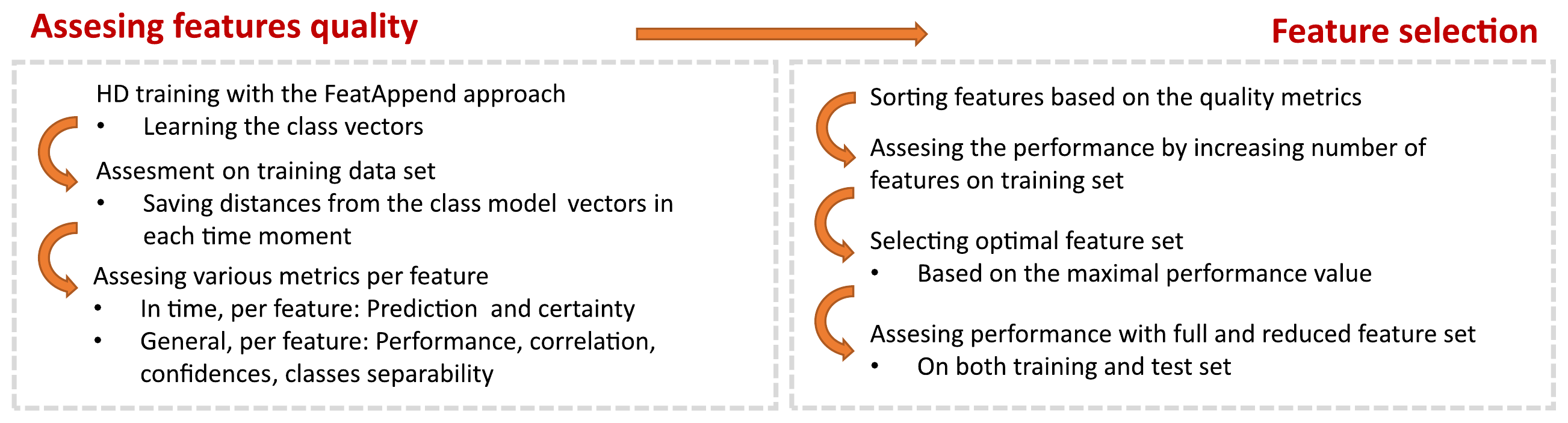} 
  \caption{Feature selection workflow with detailed steps.}
  \label{fig:FeatureSelectionWorkflow} 
  \vspace{-6mm}
\end{figure}

Metrics measured per feature on a training set can then be used to perform feature selection, as shown in Fig.~\ref{fig:FeatureSelectionWorkflow}. In this paper, we demonstrate three ways to do it. Each approach starts by ordering features based on specific quality measures: 
\begin{itemize}
    \item \textit{Feature performance}: Based on the predictions for each sample and the true labels, we can measure the performance of each feature. The exact performance metric of choice can depend on the application, and for epilepsy detection we define them later in Sec.~\ref{perfMetrics}. 
    
    \item \textit{Feature confidence}: Based on the certainty values and predictions per feature, as well as true labels per time moment, from the training set, features can be ordered based on the highest confidence. Confidence per feature is calculated as how much more certain is that feature during correct predictions ($C_{f} | TP$) versus wrong predictions ($C_{f} | FP$) and is calculated by the formula (\ref{eq_confidence2}). 

\begin{equation}\label{eq_confidence2}
Conf_f =  \frac{\overline{C_{f} | TP} - \overline{C_{f} |  FP}} {\overline{C_{f} |  FP}} 
\end{equation}    
    

    \item \textit{Feature performance and correlation}: If selection is made based only on the feature performance, it might lead to selecting features that are performing well but are highly correlated and thus maybe redundant. In this approach, we select features one by one by evaluating in each step how the performance changes when adding one of the not yet used features. This is a slower process (and computationally more complex) than sorting features based on different direct quality metrics, as in the previous two approaches. Nonetheless, it leads to a more elaborate order of features, which also includes the novelty each feature brings.
\end{itemize}

After features are ordered, the performance on the train and test set is assessed when increasing the number of features until all features are included. Prediction when using $n$ features is given by summing up distances from seizure and non-seizure model vectors of individual features as in the formula (\ref{eq_finallabel}). If the final distance is positive, the prediction (\textit{Vote}) is seizure, otherwise non-seizure. From the performance curve of the training set, the optimal number of features is chosen as the number of features giving maximal performance (or the smallest number of features without loss in the performance when compared to using all features).
In the end, as shown in Fig.~\ref{fig:FeatureSelectionWorkflow}, performance on the test set is measured for the chosen, reduced set of features. Then, we compare it with the case without feature selection.

\begin{equation}\label{eq_finallabel}
Vote =  sign \bigg[ {\sum_{i=1}^{numFeat} { \big({dist}_{NSi} - {dist}_{Si}} } \big) \bigg]
\end{equation} 

\section{Experimental setup}



\subsection{Dataset} 
For the analysis, we use the CHB-MIT database. It consists of 24 subjects with medically-resistant seizures ranging in age from 1.5 to 22 years~\cite{shoeb_application_2009, goldberger_ary_l_physiobank_2000}. It is an EEG database collected by the Children's Hospital of Boston and MIT, and contains 183 seizures overall, with an average of 7.6 $\pm$ 5.8 seizures per subject. We use the 18 channels (i.e., FP1-F7, F7-T7, T7-P7, P7-O1, FP1-F3, F3-C3, C3-P3, P3-O1, FP2-F4, F4-C4, C4-P4, P4-O2, FP2-F8, F8-T8, T8-P8, P8-O2, FZ-CZ, CZ-PZ) that are common to all patients.

The original dataset contains more than 980 hours of recordings that are divided in approximately one hour-long files. Even if the common approach in literature is using balanced data preparation, it can lead to highly overestimated performance~\cite{pale_multi-centroid_2022}. Further, training on the full dataset using HD computing is not feasible due to its complexity. Thus, as proposed in~\cite{pale_exploration_2022}, we use a data selection approach that contains all seizure segments and ten times more non-seizure data. Data is arranged in such a way that for each seizure file, seizure data is extracted and surrounded by non-seizure data, which was randomly selected from one of the files not containing any seizure. In this way, each file contains an equal ratio of seizure and non-seizure data, and can be easily used for leave-one-file-out cross-validation. 



\subsection{Feature Choices} 
In previous papers applying HD computing for epileptic seizure detection ~\cite{pale_multi-centroid_2022, pale_exploration_2022, pale_systematic_2021}, 45 features were used for classification, including mean-amplitude, 37 entropy features and 8 relative frequency domain features. Based on a literature review by~\cite{siddiqui_review_2020} discussing the importance of various features, frequency features, as well as line-length were reportedly the most useful. Thus, in this paper, we keep the mean-amplitude ($mean\_ampl$) and use both relative and absolute values of power spectral density in the five common brain wave frequency bands: delta: [0.5-4] Hz ($p\_delta$ and $p\_delta\_rel$), theta: [4-8] Hz ($p\_theta$ and $p\_theta\_rel$), alpha: [8-12] Hz ($p\_alpha$ and $p\_alpha\_rel$), beta: [12-30] Hz ($p\_beta$ and $p\_beta\_rel$), gamma: [30-45] Hz ($p\_gamma$ and $p\_gamma\_rel$), and low-frequency components: [0-0.5] Hz ($p\_dc$ and $p\_dc\_rel$) and [0.1-0.5] Hz ($p\_mov$ and $p\_mov\_rel$). To calculate the relative values, we divide the absolute values by the total power ($p\_tot$). In this case, we do not to use entropy features, as a preliminary analysis and the literature showed a very small discriminative power. In the end, we included the line-length feature ($line\_length$), 
as introduced in~\cite{esteller_line_2001}, and which showed a high discriminative power in the preliminary analysis. Thus, in total, we extract 19 features. 
Before extracting the features, the data is
filtered with a zero-phase, 4th order Butterworth band-pass filter between [1, 20] Hz. Features are extracted from data segmented into 4-second windows with a 0.5-second step. 


\subsection{HD Learning Workflow} 
The standard HD computing workflow consists of a single-pass training, where HD vectors representing different data windows (in this case, four-second windows) from the same label class are bundled all together to form a model HD vector representing that class (illustrated in Fig.~\ref{fig:HDworkflow}). This approach is simple and fast. However, all data windows are equally important, which can, in highly imbalanced datasets such as epilepsy ones, lead to the domination of more common patterns in the final vectors. As shown in~\cite{pale_exploration_2022}, this situation leads to an under-representation of less common patterns and ultimately lowers the performance. OnlineHD was proposed in~\cite{hernandez-cano_framework_2021} as an alternative, where each window vector is multiplied with a weight before being added to the model vector. The weight is defined by the similarity of the current vector to the current prototype vectors; the higher the similarity, the lower the weight, which helps in identifying more repeating patterns and lowers model saturation with them. In~\cite{pale_exploration_2022} standard HD and OnlineHD have been compared for the exact use case of epileptic seizure detection, and OnlineHD has shown to have higher performance. An essential step of OnlineHD is that, when a new sample vector is added, it is also subtracted from the opposite class if it is more similar. These steps help to make model vectors more separable and improve the performance. 

\subsection{Validation} 

\subsubsection{Feature comparison} 
A standard approach to compare features individually is the Kullback-Leibler or Jensen-Shannon divergences, which analyses the distance between the distributions of feature values for the different classes.
As explained in Sec.~\ref{sec:featureSelection} $FeatAppend$ appending approach enables various other ways to compare features. For example, it is possible to compare predictions if the model only has individual features and their certainties in time. Moreover, the final performance per feature, confidence, correlation, and class separability can be assessed too. 

\subsubsection{Performance evaluation} \label{perfMetrics}
Training and evaluation are done independently for each individual due to the subject-specific nature of epileptic seizures. For each person, we perform leave-one-seizure out cross-validation and report the performance as the average across all cross-validation runs. In the end, we report performance as average over all subjects. 

The performance of the classifier is evaluated with respect to seizure episode detection and duration-based (window-based) detection. There is an active discussion in the community on which performance measures to use to increase the interpretability of performance~\cite{ziyabari_objective_2019, shah_validation_2020}. Here we use the same metrics as in~\cite{pale_multi-centroid_2022}. More specifically, we measure sensitivity, predictivity, and F1 score on the level of episode detection, as well as on the level of episode duration. The performance at the episode level groups the signal into blocks of seizure and non-seizure. 
The performance at the duration level, on the other hand, cares about the correct prediction of each window, meaning that seizures need to be predicted correctly during their whole duration. 
Finally, to have one single final metric, we combine F1 scores for episodes ($F1E$ and $F1D$) using the geometric mean as $F1DEgmean$. 

\begin{figure}[t]
  \centering 
  \includegraphics[width=6in]{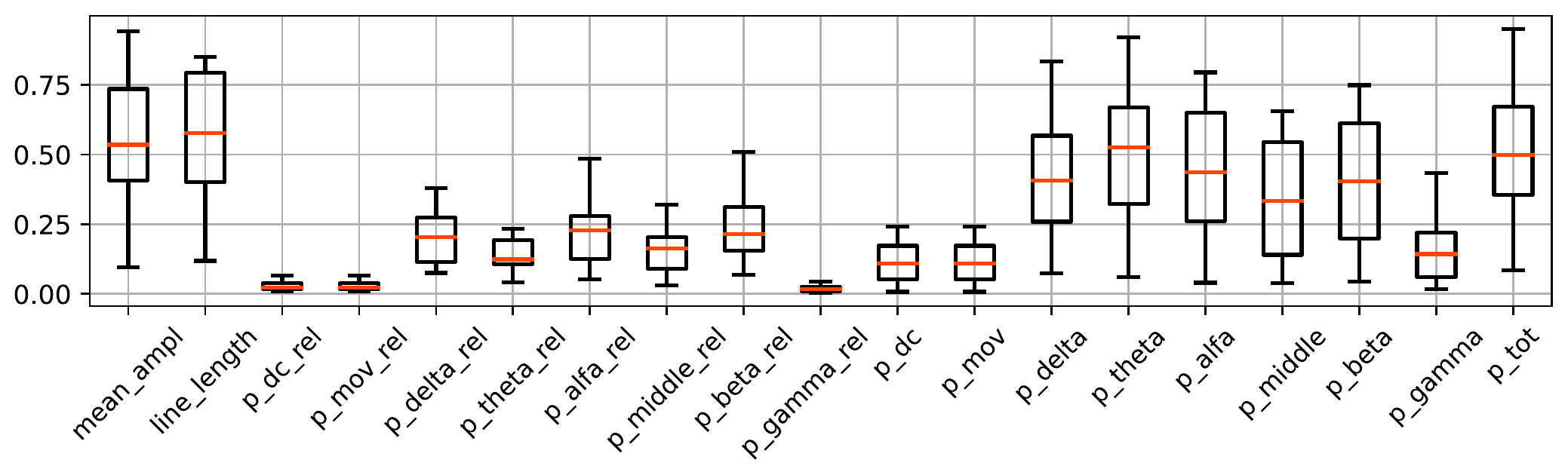} 
  \caption{Jensen-Shannon divergence of features.}
  \label{fig:FeatDivergence} 
\end{figure} 

In epilepsy detection, raw label predictions often lead to an unrealistic behavior of seizure dynamics (e.g., seizures lasting only a few seconds, or seizures that are only a few seconds apart). Thus, label post-processing is an integral part of the whole pipeline. Here we, post-process raw label predictions by performing a moving average with majority voting, using a window size of 5s.








\begin{figure}[t]
  \centering 
  \includegraphics[width=6in]{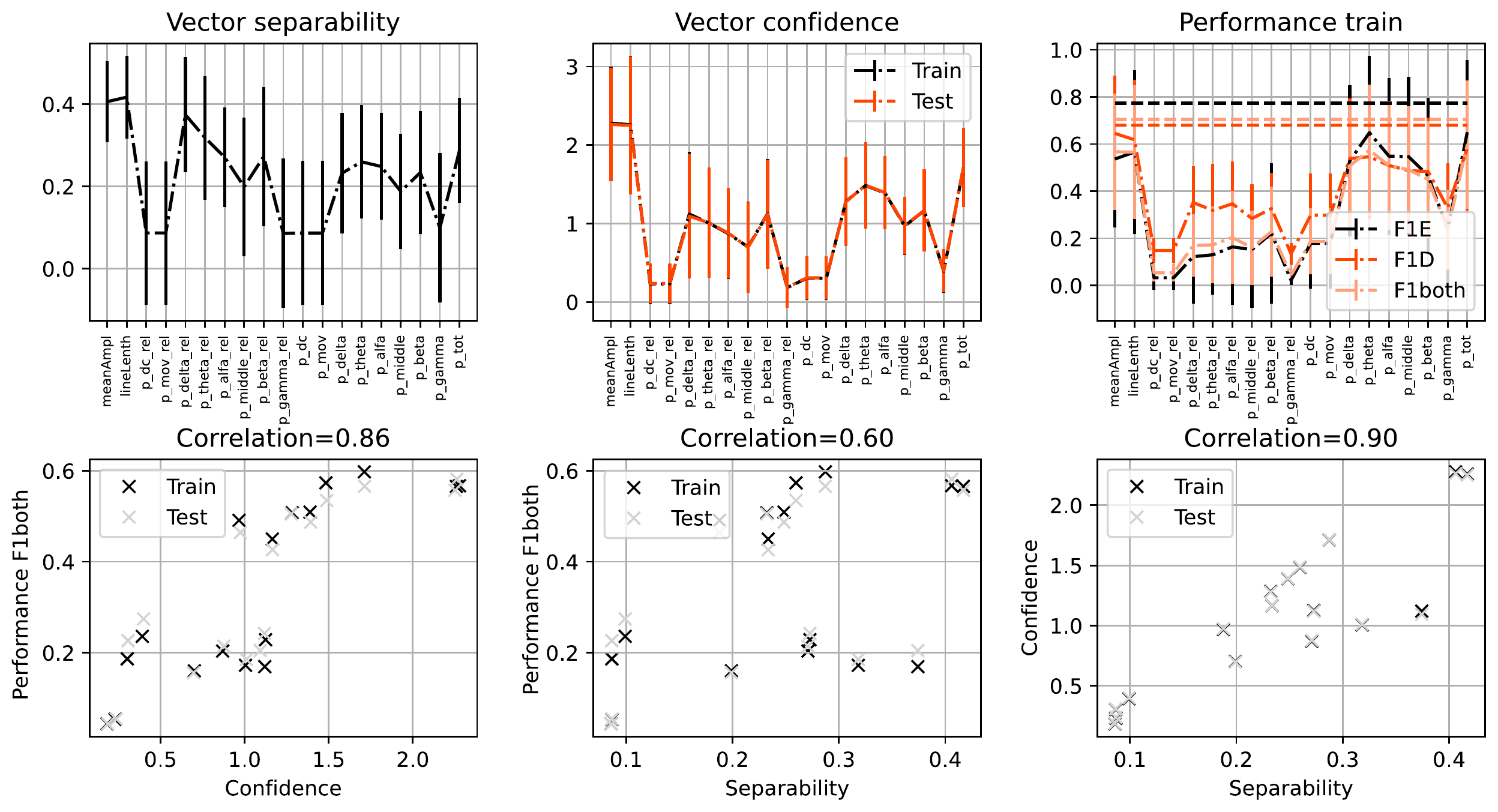} 
  \caption{Comparison of features based on $FeatAppend$ approach. The averaged values over all subjects are shown.}
  \label{fig:FeatAppendComparisonOfFeatures} \vspace{-6mm}
\end{figure} 

\section{Results} 

\subsection{Feature comparison } 

Fig.~\ref{fig:FeatDivergence} shows Jensen-Shannon divergence of 19 features we used and their distribution over all channels for all subjects. There is a clear difference between features, where mean amplitude, line length, total energy, and absolute spectral powers of the delta, theta, alpha, beta, and middle-range are quite discriminative. Relative powers seem to be less discriminative than absolute values.

Fig.~\ref{fig:FeatAppendComparisonOfFeatures} shows comparison of the features based on measures extracted using $FeatAppend$ approach. More specifically, the separability of vectors for the two classes are shown for each feature, average confidence of each feature, and the performance (F1 score for episodes, duration, and their gmean) when using only individual features. In the subplot of performance, the horizontal line shows the performance achieved when using all the features. It is visible that no single feature reaches the performance of all features, but some of them get quite close to it (i.e., mean amplitude, line length, total power, and power of delta and theta). In the bottom row, the correlation between confidence and performance, separability and performance, and separability and confidence are plotted, showing high correlation values. This confirms that the $FeatAppend$ approach can indeed be used to investigate different properties of individual features.

\begin{figure}[t]
  \centering 
  \includegraphics[width=6in]{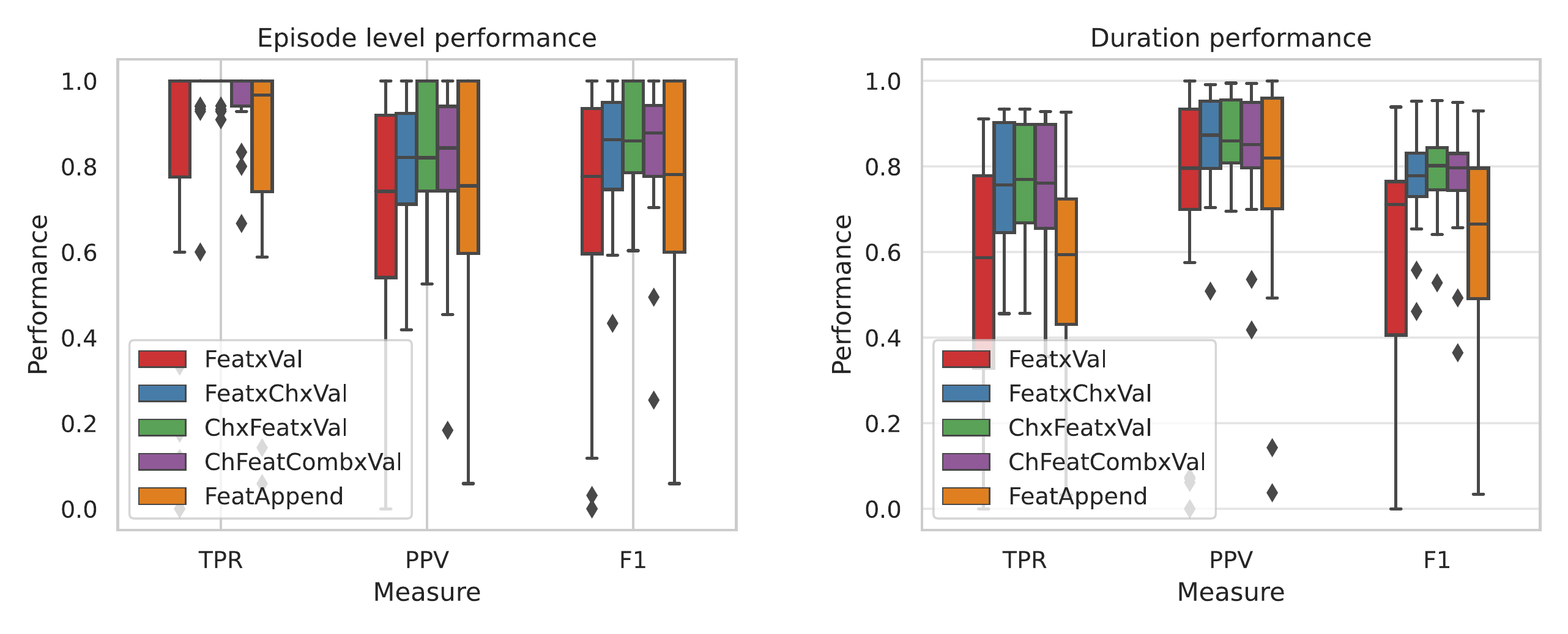} 
  \caption{Performance of different EEG encoding approaches.}
  \label{fig:EncodingPerformanceComparison}
  \vspace{-6mm}
\end{figure}

\subsection{Encoding comparison} 

Next, as $FeatAppend$ is one of the possible approaches to encode all EEG data aspects into the vectors to use them for HD learning, we compare it with other possible encoding approaches. Fig.~\ref{fig:EncodingPerformanceComparison} shows the average performance for all subjects without any post-processing. Performances on the level of seizure episodes and seizure duration are shown. TPR or true positive ratio is the measure of sensitivity, PPV or positive predictive value is the measure of precision, and F1 score is gmean between TPR and PPV. 

$Feat\times Val$ encoding, which does not takes channel information into account,  leads to lower performance encoding types that take channel information into account. There is no significant difference in performance between the three approaches that include channel information: $Feat \times Ch \times Val$, $Ch\times Feat\times Val$ and $ChFeatComb\times Val$. 
The $FeatAppend$ approach, even if including channel information, yields a smaller performance than the three approaches that utilize channel information, probably due to the smaller number of dimensions per feature. Yet, performance is not worse than the $Feat\times Val$ approach. 

\begin{figure}[t]
  \centering 
  \includegraphics[width=6in]{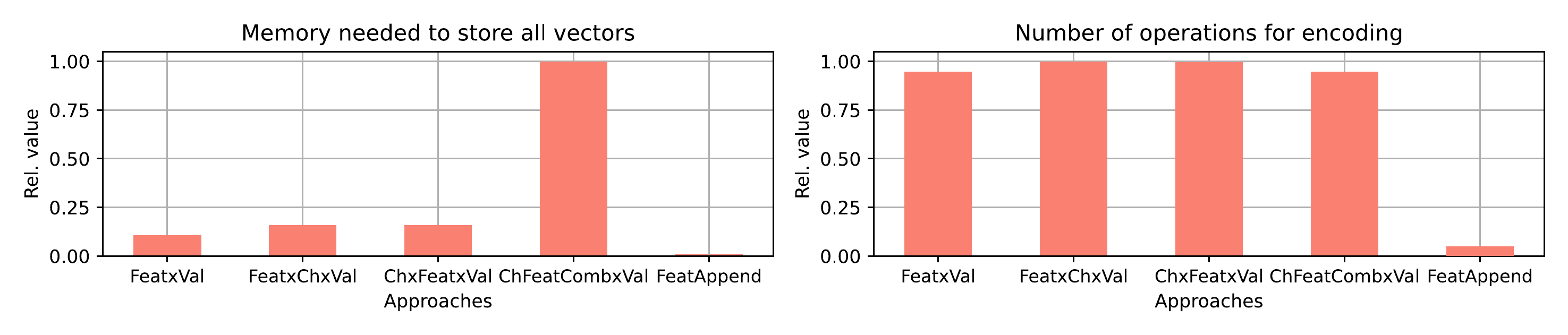} 
  \caption{Comparison of encoding approaches in terms of memory needed and number of binding and bundling operations needed to encode one data window.}
  \label{fig:MemoryAndTimeForEncoding} 
  \vspace{-8mm}
\end{figure}

Fig.~\ref{fig:MemoryAndTimeForEncoding} shows the memory needed to store all HD vectors (for channels, features, and values) for each of the approaches is shown. Relative ratios are shown as memory directly depends on the chosen dimension D (in this case 19000) and whether HD vectors are binary or not. Due to the large number of combinations of features and channels, the $ChFeatComb\times Val$ approach is the most memory-demanding one, and scales the worst. $FeatAppend$ approach is the approach that requires the least amount of memory as base vectors have lower $D/numFeat$ dimensions. 

Further, the relative number of bundling and binding operations needed to encode one window of data to the HD vector representing it is shown as well in Fig.~\ref{fig:MemoryAndTimeForEncoding}. The $FeatAppend$ approach requires significantly less operations due to the effectively smaller number of dimensions per vector for each feature. From other approaches, $ChFeatComb\times Val$ and $Feat\times Val$ require slightly less operations then $Feat \times Ch \times Val$ anr $Ch\times Feat\times Val$.

\subsection{Feature selection} 

\begin{figure}[t]
  \centering 
  \includegraphics[width=6in]{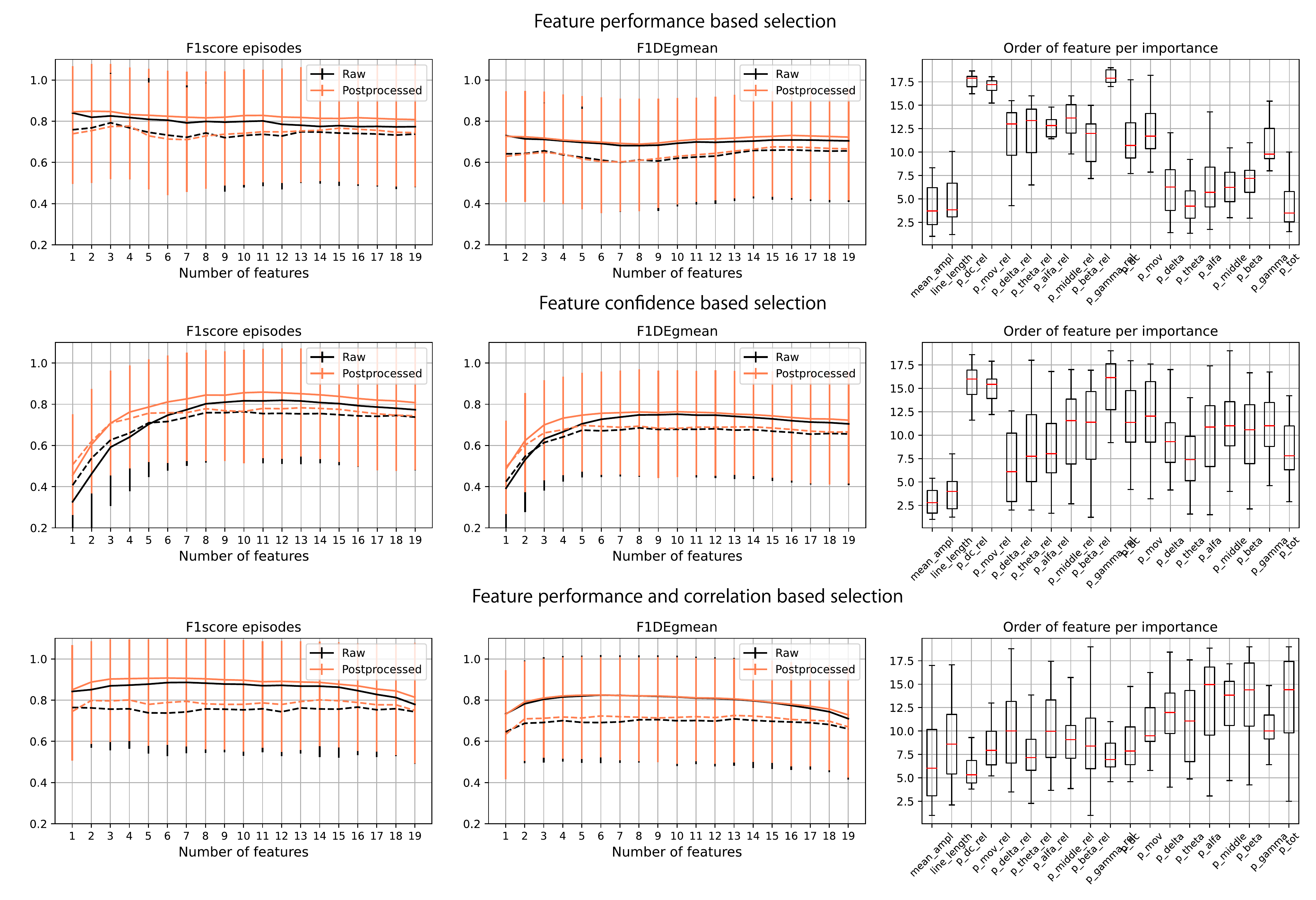} 
  \caption{Performance evolution by incrementally adding one-by-one new feature for three approaches: based on feature performance, feature confidence and both feature performance and correlation.}
  \label{fig:PerformanceCurves} 
  \vspace{-8mm}
\end{figure}

Fig.~\ref{fig:PerformanceCurves} shows the performance when adding features one-by-one, for each of the three methods described in Sec.~\ref{fig:PerformanceCurves}. The difference between approaches relies in how features are ordered. In the first row of Fig.~\ref{fig:PerformanceCurves} features are ordered based on their performance metric (in this case, their F1DE performance) and in the second row based on average feature confidences given by equation (\ref{eq_confidence2}). Finally, in the third row, features are added one by one choosing the best feature to increase performance when added to previously chosen features. This approach can be referred to as more optimal order and choice of features, as it takes into account correlation between features. In this case, performance increases and reaches performance higher than when using the first two approaches.  

The last column shows the boxplots of feature orders for each feature selection method, for all subjects. The smaller the ranking number, the sooner that feature was chosen in the incremental feature selection. What can be noticed is that the ranking of features in the first two approaches when using solely feature performance or feature confidence is similar to the results given by the discriminative power analysis shown in Fig.~\ref{fig:FeatDivergence}. In the last case of more optimal feature selection, feature order is slightly different due to feature correlations that were taken into account.

\begin{figure}[t]
  \centering 
  \includegraphics[width=6in]{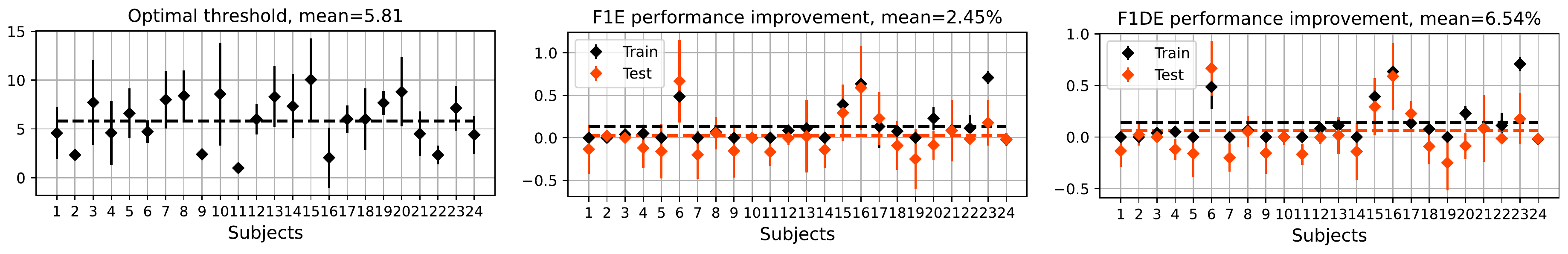} 
  \caption{Optimal number of features and performance after feature selection. Example is showed for feature selection using both feature performances and correlations. The horizontal lines represent the average for all subjects.}
  \label{fig:ExampleFeatSelectionRes} 
  \vspace{-10mm}
\end{figure}

Fig.~\ref{fig:ExampleFeatSelectionRes} shows the results of optimal feature selection for every subject. The first graph shows the chosen number of features per subject and average for all subjects, which is shown with the horizontal line. The following two graphs show the performance improvement (or decrease), i.e., F1 for episodes and gmean of F1 for episodes and duration, for the training and testing set. In the title of the figures, we provide the average performance for all subjects on the test set. A significant variability exists between the number of features chosen per subject, ranging from 1 to 10 features, with an average of 5.8 or 30\% of features. 

In Table~\ref{tab:PerformanceResults} shows more detailed results. More precisely, results for all three feature selection methods are given. For each of the methods, we show results when we optimized the F1 score only for episodes (F1E), or when the F1 score for seizure duration is also taken into account (F1DE). Further, in the table, average F1 and F1DE performances for both train and test are shown before and after label post-processing. 
What can be noticed is that when optimizing only F1E, fewer features are needed, but it usually leads to a smaller performance increase for F1DE. When F1DE is optimized, this leads to a significant performance increase both for F1DE and F1E but at the price of a slightly higher number of features chosen. In general, the performance increase is smaller on the test set than on the train set, which is reasonable as the optimal number of features was chosen based on the train set without knowledge about the test set. Yet, the performance increase is not negligible, ranging up to 7\% for the test set. 

When comparing the three different feature selection approaches, there are slight differences in the optimal number of features chosen and the performance gained. Feature selection based only on the performance or confidence leads to a slightly higher number of features but not a higher performance increase. This is due to information about feature correlations that are missing.

Finally, the code and data required to reproduce the
presented results are available online as open-source (link will be available after the double-blind peer-review process).

\begin{table}[t]
  \centering 
  \caption{Optimal number of features and performance change for different feature selection approaches.}
  \resizebox{\textwidth}{!}{\begin{tabular}{lllllllllll}
  \toprule
    \textbf{Feat. selection} & \textbf{Perf.} & \textbf{Nr.} & \multicolumn{2}{c}{\textbf{Train Raw}} &  \multicolumn{2}{c}{\textbf{Train Post}} &  \multicolumn{2}{c}{\textbf{Test Raw}} &  \multicolumn{2}{c}{\textbf{Test Post}}\\
    \textbf{approach} & \textbf{used} & \textbf{Feat.} &\textbf{F1E} & \textbf{F1DE} & \textbf{F1E} & \textbf{F1DE} & \textbf{F1E} & \textbf{F1DE} & \textbf{F1E} & \textbf{F1DE}\\
    \midrule
    \textbf{Feature} & F1E & 2.84 & 12.29 & 4.63 & 6.75 & 1.43 & 4.78 & -1.40 & -0.27 & -4.53\\ 
    \textbf{performance} & F1DE & 6.99 & 9.41 & 7.49 & 6.60 & 5.39 & 4.94 & 3.69 & 3.64 & 1.90\\ 
    \midrule
    \textbf{Feature}  & F1E &5.98 & 10.92 & 8.55 & 6.59 & 5.56 & 4.00 & 3.75 & 2.90 & 2.28 \\ 
    \textbf{confidence} & F1DE & 6.62 & 9.44 & 10.26 & 6.60 & 7.81 & 3.56 & 6.03 & 3.43 & 4.82\\
    \midrule
    \textbf{Feat}  & F1E & 2.65 & 14.61 & 8.91 & 9.35 & 5.45 & 5.84 & 1.88 & 1.79 & -1.06 \\ 
    \textbf{perf and corr} & F1DE & 5.81 & 13.29 & 14.07 & 9.44 & 10.84 & 2.45 & 6.54 & 5.27 & 6.96 \\
    \bottomrule
  \end{tabular}}
  \label{tab:PerformanceResults} 
  \vspace{-5mm}
\end{table}

\vspace{-3mm}
\section{Discussion} 
This paper draws attention to not yet discussed and properly explored topics of mapping and encoding spatio-temporal data such as EEG or EMG to HD vectors. Although HD computing has been utilized and has shown promising results for various biomedical applications (in particular utilizing EEG and EMG), most works in the literature only use raw data (or LBP values), namely, only one feature per channel. Thus, the optimal encoding when more features per channel or more available data modalities are used remained unclear. 

In this work, we propose five alternatives to encode feature values for all channels of one data window into an HD vector and test it on epileptic seizure detection. Our results show that including channel information is beneficial for epileptic detection performance, but that the order in which features and channels are mapped to corresponding values is not relevant. Further, we show that the $ChFeatComb\times Val$ approach is the most memory demanding and that $Feat \times Ch \times Val$,  $ChxFeat\times Val$ and  $FeatAppend$ are comparable and more appropriate. $FeatAppend$ requires the least amount of memory and operations to encode vectors due to the effectively smaller number of dimensions per feature vector.
By creating individual vectors for each feature and appending them next to each other, in $FeatAppend$ performance is slightly reduced. This situation is probably due to fewer HD dimensions used per feature. Thus, it could be improved by increasing the total dimension of vectors, but this also leads to higher memory requirements, which could make it less friendly for wearable devices. 

By utilizing the $FeatAppend$ approach, we present, to the best of our knowledge, for the first time in the literature, a way to perform feature selection using HD computing. An incremental feature approach was tested with three different methods to determine the order of features to be added. All approaches led to a significant reduction of features while keeping or even significantly improving the performance compared to using all features. In the future, approaches using feature elimination could be similarly tested. 

$FeatAppend$ is interesting not only from a feature selection perspective but also from the clinical perspective of a deeper understanding of various features and their properties. For example, we investigated several measures per feature: performance, probabilities of decisions, confidences, correlation, and separability of classes, which can all lead to knowledge discovery related to the usefulness of features. 
This approach can be adapted to $ChannelAppend$ and be used in an analogous way for channel comparison, channel selection, and potentially seizure localization.  

Thus we hope that in the future, this work will serve as inspiration for further research and novel ideas in the direction of feature exploration, feature and prediction interpretability, and channel selection. More specifically, for epileptic seizure detection, seizure localization and more detailed feature quality assessment for different seizure-type classifications could be research venues. 


\paragraph{Limitations}
As the majority of works dealing with large and unbalanced databases, such as epilepsy ones, this work was done only on a sub-selection of the of CHB-MIT database due to computational limitations. Namely, training on the whole database  
takes an extensive amount of time. Thus, by using a random selection of non-seizure data and 10 times more non-seizure than seizure data, we tried to minimize the negative effects of not using the whole database. Still, in the future, it would be interesting to optimize and adapt the code for GPU processors and reproduce the analysis with the whole database. 

The encoding approach based on feature appending is simple and enables a detailed comparison of each feature, but it has the drawback that the dimension of the HD vectors scales with the number of features, potentially making it more memory and time-consuming. At the same time, reducing the number of dimensions per feature lowers the information capacity that can be stored and ultimately lowers the performance. Thus, different approaches that enable the comparison of a high number of features without limiting the number of dimensions per feature should be designed in the future. 

\vspace{-3mm}
\section{Conclusion} 
In this paper, we have demonstrated how a novel approach of hyperdimensional computing (HD) can be used as an alternative to standard state-of-the-art ML approaches in the use case of epileptic seizure detection. We explored the not yet addressed topic of optimal encoding of spatio-temporal data, such as EEG, and all the information it entails, into the HD vectors. We compared different approaches with respect to their memory and computational complexity, as these metrics are of high interest for wearable devices for continuous monitoring of diseases, such as epilepsy or stress conditions. Hence, those approaches with lower complexity could make longer battery lifetimes feasible and make a step towards a preventive healthcare. 

In addition, we have demonstrated how HD computing framework can be utilized to perform feature selection by choosing an adequate encoding. Three approaches were tested and led to a significant reduction of features, while keeping or even significantly improving the performance compared to using all the features.
Overall, we expect that this work can serve as inspiration for further research and novel ideas in the direction of feature exploration, feature and prediction interpretability, and channel selection.



\bibliography{references.bib}



\end{document}